\documentclass[runningheads,a4paper]{llncs}

\usepackage{amssymb}
\usepackage{graphicx}
\usepackage{times}
\usepackage{url}
\usepackage{latexsym}

\urldef{\mailsa}\path|qi.li@pku.edu.cn, lts_417@hotmail.com, chbb@pku.edu.cn|    
\newcommand{\keywords}[1]{\par\addvspace\baselineskip
\noindent\keywordname\enspace\ignorespaces#1}

\begin{document}

\mainmatter  % start of an individual contribution
\title{Learning Word Sense Embeddings from Word Sense Definitions}

\author{Qi Li\and Tianshi Li\and Baobao Chang}

\institute{Key Laboratory of Computational Linguistics, Ministry of Education \\
School of Electronics Engineering and Computer Science, Peking University \\
No.5 Yiheyuan Road, Haidian District, Beijing, 100871, China\\
Collaborative Innovation Center for Language Ability, Xuzhou, 221009, China \\
\mailsa
}

\maketitle

\begin{abstract}

Word embeddings play a significant role in many modern NLP systems. Since learning one representation per word is problematic for polysemous words and homonymous words, researchers propose to use one embedding per word sense. Their approaches mainly train word sense embeddings on a corpus. In this paper, we propose to use word sense definitions to learn one embedding per word sense. Experimental results on word similarity tasks and a word sense disambiguation task show that word sense embeddings produced by our approach are of high quality.
\end{abstract}

\section{Introduction}
With the development of the Internet and computational efficiency of processors, gigantic unannotated corpora can be obtained and utilized for natural language processing (NLP) tasks. Those corpora can be used to train distributed word representations (i.e. word embeddings) which play an important role in most state-of-the-art NLP neural network models. The word embeddings capture syntactic and semantic properties which can be exposed directly in tasks such as analogical reasoning \cite{Mikolov2013DistributedRO}, word similarity \cite{HuangEtAl2012} etc. Prevalent word embedding learning models include Skip-gram \cite{Mikolov2013DistributedRO}, Glove \cite{Pennington2014GloveGV} and variants of them.

Basic Skip-gram \cite{Mikolov2013DistributedRO} and Glove \cite{Pennington2014GloveGV} output one vector for each word.  However, multi-sense words (including polysemous words and homonymous words) should inherently have different embeddings for different senses. Therefore researchers propose to use one embedding per word sense \cite{Reisinger2010MultiPrototypeVM,HuangEtAl2012,Chen2014AUM,Neelakantan2014EfficientNE,Tian2014APM,Iacobacci2015SensEmbedLS,Liu2015TopicalWE,wu2015sense,Li2015DoME}. Previous work tends to perform word sense induction (WSI) or word sense disambiguation (WSD) on the corpus to determine the senses of words. Then they train the word sense embeddings on it using variants of Skip-gram or other approaches. However, the result of WSI or WSD on the corpus is not reliable and the errors from WSI or WSD will have bad effect on the quality of word sense embeddings. Besides, these approaches normally produce bad embeddings for rare word senses.

Lexical ontologies such as WordNet \cite{Miller1992WordNetAL} and BabelNet \cite{Navigli2012BabelNetTA}  are built by specialists in linguistics and they provide semantic information of word senses including their definitions. Different from determining word senses by WSI or WSD models, semantic information provided by lexical ontologies is normally accurate and reliable. To utilize the accurate information of word senses provided by lexical ontologies, we propose an approach based on recurrent neural networks (RNN) to learn word sense embeddings from word sense definitions. Our approach learns both word sense embeddings and a definition understanding model. Since the collection of definitions is much smaller in scale than a corpus for embedding training, our approach is less time-consuming comparing with corpus-based learning approaches. Experimental results show that the word sense embeddings are of high quality for both common and rare words and the definition understanding model can understand other natural language text besides word sense definitions.

Our contributions can be summarized as follows:

 \begin{itemize}
\item We propose to learn word sense embeddings from word sense definitions using RNN-based models.
\item Different from previous embedding learning approaches, our learning is conducted in a supervised paradigm. 
\item Our approach is less time-consuming comparing with corpus-based learning approaches. 
\item Our approach treats senses of rare words and of common words equally since definitions have no tendency to common words and this is hard to achieve for corpus-based learning approaches. 
\end{itemize}

The rest of this paper is organized as follows:  Section \ref{approach} presents details of our approach. Section \ref{experiment} reports experimental results. Section \ref{relate} introduces the related work.  Section \ref{conclusion} concludes our work.

\section{Methodology}
\label{approach}

While a corpus presents distributional properties of words, definitions provide semantic information of word senses in a compositional way. Therefore, we believe that we can compute word sense embeddings from definitions. We choose to use recurrent neural networks to model semantic compositionality because RNN-based models have been shown to be able to model semantic compositionality in many tasks, such as neural machine translation \cite{Bahdanau2014NeuralMT,Hermann2015TeachingMT,Kalchbrenner2013RecurrentCT},  text entailment recognition \cite{Rocktschel2015ReasoningAE} etc. 

\subsection{Definition Understanding Model}
\begin{figure}
\centering
\includegraphics[width=12.4cm]{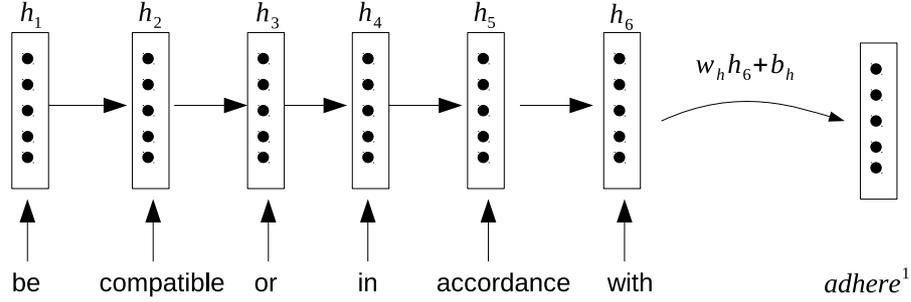}
\caption{Mapping definitions to word sense embeddings with RNN-based model. }
\label{structure}
\end{figure}

A word sense definition is a word sequence: $\{x_1,x_2,...,x_n\}$. As Figure \ref{structure} shows, RNN models take word embeddings of the words in definitions one by one and update the internal memory according to its computation unit. The output of the RNN at the last word of the definition, i.e., $h_n$ is assumed to contain the semantic meaning of the definition. Hence we map $h_n$ to sense embedding space with a transformation matrix:

\begin{equation}
\tilde{e_{ws}}= W_hh_n+b_h
\end{equation}
where $W_h$ is the transformation matrix, $b_n$ is the bias term and $\tilde{e_{ws}}$ is the sense embedding computed by the definition understanding model.

The specific RNN model can be a vanilla RNN model, Gated Recurrent Unit (GRU) \cite{Chung2014EmpiricalEO} or Long Short Term Memory (LSTM) \cite{Hochreiter1997LongSM}. Comparing with vanilla RNNs, LSTMs and GRUs can hold long-term information, i.e., they can alleviate the gradient vanishing and information-forgetting problem associated with the vanilla RNN for long sequences \cite{Chung2014EmpiricalEO,Hochreiter1997LongSM}.

\subsection{Training Definition Understanding Model with Definitions of Monosemous Words}
Having determined the model structure, the challenge is how to train the RNN-based definition understanding model. Since the contexts of a monosemous word are associated with its only word sense, we assume its word sense embedding is similar to its word embedding. So we initialize sense embeddings of monosemous words with their word embeddings trained with Skip-gram \cite{Mikolov2013DistributedRO} on a corpus and thus we can train the RNN-based model parameters with sense embeddings of monosemous words as target and their definitions as inputs. The words in definitions are represented by their word embeddings. Word embeddings are kept fixed during the whole training process. As word embeddings trained on a corpus provide distributional properties of the words, our approach provides a supervised training for model parameters by combining distributional and compositional properties of word senses. The objective function of this training step is
\begin{equation}
J_1= -\sum_{w\in V_{mono}}cos(e_{ws},\tilde{e_{ws}})
\label{j1}
\end{equation}
where $V_{mono}$ is the set of monosemous words and $e_{ws}$ is the initialized sense embedding of the monosemous word $w$ and $\tilde{e_{ws}}$ is the word sense embedding produced by our RNN-based definition understanding model. With this objective function, we train $\tilde{e_{ws}}$ to be similar to $e_{ws}$. Both the sense embeddings and word embeddings used in definitions are fixed in this step, i.e., we only train model parameters in this step.

\subsection{Word Sense Embedding Learning}
We have trained RNN-based definition understanding model with sense embeddings of  monosemous words and their definitions in the previous step. However, we still haven't use definitions of senses of multi-sense words. Since the word embeddings are trained according to the co-occurrences from a corpus, the word embedding of a multi-sense word usually represent the most common sense better and this is shown by our nearest neighbor evaluation in Section \ref{knn}. So it is not appropriate to initialize all sense embeddings of a multi-sense word with its word embedding. Since WordNet provides a set of synonyms for each word sense, we initialize the embedding of a word sense using the word embedding of a synonym which contains only one sense. If there is no synonym conforms to this condition, we initialize the sense embedding with the word embedding of a word in its definition which has the largest cosine similarity with the original word. Besides, the similarity should exceed a threshold $\delta$ which we set to be 0.2 or we will use the word embedding of the target word sense to initialize its sense embedding. Although these sense embeddings are simply initialized, they still contain meaningful semantic information for training the definition understanding model and reversely they will also be tuned by our model. Comparing with the last step, we still optimize the cosine similarities between embeddings produced by the RNN-based definition understanding model and the initialized word sense embeddings. The difference is that in this step we use definitions of both monosemous words and multi-sense words and update all sense embeddings jointly including sense embeddings of monosemous words because some of them are of low quality if the words are of low frequency in the corpus their word embeddings are trained. The objective function is as follows:
\begin{equation}
J_2= -\sum_{w\in V}\sum_{s\in S_{w}}cos(e_{ws},\tilde{e_{ws}})
\label{j2}
\end{equation}
where $V$ is the whole word set and $S_{w}$ is the set of word senses of word $w$. To sum up, we make word sense embeddings and RNN-based definition understanding model tune each other in this step.

\subsection{Training with Word Sense Embeddings to Represent Words in Definitions}
\label{lastphase}
In the previous two steps, we train the definition understanding model and learn sense embeddings jointly using word embeddings to represent words in definitions. However, some words in definitions are multi-sense words and therefore to use sense embeddings to represent those words is assumed to be more appropriate. Besides, it can also be seen as an application of the word sense embeddings trained in the last step.

To this end, we perform WSD for the words in definitions. We apply S2C (simple to complex) strategy described in \cite{Chen2014AUM} to implement WSD. Specifically, we identify the senses for words with less senses first and then for words with more senses. We compute the cosine similarity between each sense embedding of a word with its context embedding and choose the sense with greatest cosine similarity with the context embedding as the sense of the word. The context embedding is the average embedding of some other words in definitions. These words include nouns, verbs, adjectives and adverbs. We use the sense embeddings of those words whose senses have been identified and use the word embeddings of the rest words. 

The objective function is the same as the previous step, but we use sense embeddings to represent words in definitions in this step and their sense embeddings are updated for the optimization of the objective function.

\section{Experiments}
\label{experiment}
We present qualitative evaluations and quantitative evaluations in this section. To show our word sense embeddings capture the semantics of word senses, we present the nearest neighbors of a word sense based on cosine similarity between the embedding of the center word sense and embeddings of other word senses. Besides, to show our model can actually understand a definition or any description, we present the most matched word senses for a given description according to our RNN-based definition understanding model. In our quantitative evaluations, we evaluate the sense embeddings on word similarity tasks and a word sense disambiguation task.

\subsection{Setup}
 We use WordNet 3.0\footnote{http://wordnet.princeton.edu/} as the lexical ontology to acquire the definitions of word senses. We choose the publicly released 300 dimensional vectors\footnote{https://code.google.com/archive/p/word2vec/} trained with Skip-gram\cite{Mikolov2013DistributedRO} on part of Google News dataset (about 100B words) as word embeddings used in our approach. We also take the word embeddings as our baseline. We randomly initialize model parameters within (-0.012, 0.012) except that bias terms are initialized as zero vectors. We adopt Adadelta \cite{Zeiler2012ADADELTAAA} with mini-batch to minimize our objective functions and set the initial learning rate to be 0.12.

\subsection{Qualitative Evaluations}
\label{knn}
To illustrate the quality of our word sense embeddings, we show the nearest neighbors of words and of their senses in Table \ref{kneighbor}. The nearest neighbors of words are computed using word embeddings. The nearest neighbors of word senses are computed using word sense embeddings trained with our definition understanding model which uses GRU as its specific RNN. We leave out the sense numbers of nearest senses because the numbers are meaningless to be presented here. As can be seen, the nearest neighbors of the center words are normally associated with the most common sense of the word. Whereas, the nearest neighbors of word senses are associated with the corresponding sense of the word. Besides, some nearest neighbours (e.g., ''supernova'', ''dainty'') are rare to be seen in a corpus. Therefore it indicates even the sense embeddings of rare words are meaningful.
\begin{table*}
\centering
\begin{tabular}{|l|l|}
  \hline
Center Word/Sense&Nearest Neighbors\\  \hline\hline
bank&ATM machines, Iberiabank, automated teller machines\\  \hline
$bank1$&financial, deposit, ATMs\\  \hline
$bank2$&riverbank, water,  slope \\  \hline\hline
star&matinee idol, singer, superstar\\  \hline
$star1$&asteroid, celestial, supernova\\  \hline
$star2$&legend, standout, footballer\\  \hline\hline
pretty&wonderfully, unbelievably, nice\\  \hline
$pretty1$&remarkably, extremely, obviously\\  \hline
$pretty2$&beauteous, dainty, lovely\\  \hline\hline
  \hline
\end{tabular}
\caption{Nearest neighbors based on cosine similarity between word embeddings or sense embeddings.}
\label{kneighbor}
\end{table*}

Although we train RNN-based definition understanding model to map a definition to its sense embedding, we will show the RNN-based definition understanding model can also understand descriptions we made up. We compute the cosine similarities between the embedding produced by our model according to the description and all word sense embeddings to find those most matched word senses. We still choose GRU as the specific RNN. Table \ref{kneighbordef} shows the most matched word senses to the given descriptions.  The descriptions in the upper subfield are definitions from WordNet and those in the under subfield are casual descriptions made up by us. Most of the predicted words match the meaning the descriptions convey and those that don't exactly match (e.g., ''gullible'', ''falsifiable'') are semantically relevant. The predicted words of the descriptions we make up are coincident with the descriptions. That illustrates our definition understanding model is effective to understand natural language.

\begin{table*}
\centering
\begin{tabular}{|p{8cm}|l|}
  \hline
 Description & Most Matched Words\\  \hline\hline
free of deceit&aboveboard, gullible, genuine\\  \hline
 causing one to believe the truth of something&prove, convince, falsifiable \\  \hline
 make (someone) agree, understand, or realize the truth or validity of something&convince, inform, acknowledge \\  \hline\hline
 the place where people live in&home, dwellings, inhabited\\  \hline
a machine we use every day&counter, computer, dishwasher\\  \hline
the animal which lives in the sea&clam, nautilus, stonefish\\  \hline
  \hline
\end{tabular}
\caption{Using our definition understanding model to find the most matched word senses for descriptions.}
\label{kneighbordef}
\end{table*}

\subsection{Quantitative Evaluations}

\subsubsection{Word Similarity Evaluation on WordSim-353}

WordSim-353 dataset \cite{Finkelstein2001PlacingSI} consists of 353 pairs of nouns which are associated with human judgments on their similarities without context information. The evaluation metrics on this dataset is the Spearman's rank correlation coefficient $\rho$ between the average human score and the cosine similarity scores predicted by the system. 

Following \cite{HuangEtAl2012,Iacobacci2015SensEmbedLS,Reisinger2010MultiPrototypeVM}, we use weighted average of cosine similarities between each possible word sense pair as the similarity of the two words. Since there is no context provided, the weights can be uniformly distributed which is adopted by \cite{HuangEtAl2012,Reisinger2010MultiPrototypeVM} or be determined by word sense frequency in the training set which is adopted by \cite{Iacobacci2015SensEmbedLS}. We choose to take the weights uniformly distributed. The following equation describe the weighted strategy:

\begin{eqnarray}
&&WeiSim(w,w') = \sum_i^{n_1}\sum_j^{n_2} p(s_i|w)p(s_j|w')cos(e_{w_{si}}, e_{w_{sj}'})
\end{eqnarray}
where $w$ and $w'$ are the two given words, $n_1$ and $n_2$ are the number of senses of the two words and $p(s_i|w)$ and $p(s_j|w')$ are the normalized weights to use $s_i$ and $s_j$ to compute similarity, $e_{w_{si}}$ and $e_{w_{sj}'}$ are sense embeddings. 

\begin{table*}
\centering
\begin{tabular}{|l|l|}
\hline
 \hline System&$\rho\times100$\\
  \hline
Reisinger and Mooney \cite{Reisinger2010MultiPrototypeVM} tf-idf (Wiki2.05B)&76.0\\
Huang et al. \cite{HuangEtAl2012} (0.99B)&71.3\\
Neelakantan et al. \cite{Neelakantan2014EfficientNE}  (0.99B)&71.2\\
Wu and Giles \cite{wu2015sense}&73.9\\
Iacobacci et al. \cite{Iacobacci2015SensEmbedLS} &\textbf{77.9}\\
\hline
SG (100B)&66.5\\
SG (100B)+Def (Word) (vanilla RNN)&67.4(+0.9)*\\
SG (100B)+Def (Word)+Def (Sense) (vanilla RNN)&68.2(+0.7)**\\
SG (100B)+Def (Word) (LSTM)&74.1(+7.6)*\\
SG (100B)+Def (Word)+Def (Sense) (LSTM)&\textbf{75.0}(+0.9)**\\
SG (100B)+Def (Word) (GRU)&73.7(+7.2)*\\
SG (100B)+Def (Word)+Def (Sense) (GRU) &74.7(+1.0)** \\
\hline
\end{tabular}
\caption{Performances on WordSim-353. The bottom subfield shows the performance of different settings of our system. SG represents just using word embeddings we acquired. Def (Word) represents the step in which we use word embeddings to represent words in definitions to train model and sense embeddings.  Def (Sense) represents the step in which we use sense embeddings to represent words in definitions to train model and sense embeddings. * indicates statistical significant differences in t-test between performances of SG(100B)  and SG (100B)+Def (Word). ** indicates statistical significant differences in t-test between performances of SG (100B)+Def (Word)  and SG (100B)+Def (Word)+Def (Sense) with the same RNN model. }
\label{noncontextwordsimilarity}
\end{table*}

Table \ref{noncontextwordsimilarity} shows our results compared with previous approaches. Reisinger and Mooney \cite{Reisinger2010MultiPrototypeVM} propose to cluster the contexts of each word into groups and make each cluster a distinct prototype vector. Huang et al. \cite{HuangEtAl2012} also use contexts to determine the number of senses of a word and use global context to improve word representations. Neelakantan et al. \cite{Neelakantan2014EfficientNE}  extend Skip-gram \cite{Mikolov2013DistributedRO} to learn multiple embeddings per word. Wu and Giles \cite{wu2015sense} cluster word senses and learn word sense embeddings from related Wikipedia concepts. Iacobacci et al. \cite{Iacobacci2015SensEmbedLS} use BabelNet \cite{Navigli2012BabelNetTA} as the word sense inventory and apply WSD to a corpus before they train word sense embeddings with Continuous Bag of Words (CBOW) architecture \cite{Mikolov2013EfficientEO}. 

As can be seen, our approach achieves significant improvement over the original word embeddings we use.  Most improvements come from the step we train word sense embeddings with our RNN-based models when the definitions are still represented by word embeddings. We achieve further significant improvements when we continue to jointly train the model and learn sense embeddings using sense embeddings trained in the previous step to represent words in definitions. It can be seen as an application of sense embeddings in a natural language understanding task, so it also illustrates our sense embeddings are better than word embeddings for natural language understanding from the perspective of real-world natural language understanding tasks. The LSTM version and GRU version present comparable performances to be used as the specific RNN in definition understanding model and vanilla RNN performs much worse than the other two models. This is in accordance with what previous work illustrated about the superiority of GRUs and LSTMs over vanilla RNNs \cite{Chung2014EmpiricalEO,Hochreiter1997LongSM}.

\subsubsection{Word Similarity Evaluation on Stanford's Contextual Word Similarities}
Since we need a context to determine the sense of a word when we use the sense embeddings in real-world tasks and evaluation on context-free word similarity datasets does not allow us to determine the sense, it cannot fully reveal the quality of our sense embeddings. Stanford's Contextual Word Similarities (SCWS) \cite{HuangEtAl2012} is a data set which provides the contexts of the target words. The way we determine the sense of the target words is the same S2C strategy we described in Section \ref{lastphase}. Having determined the senses of the target words,  we compute cosine similarity of their sense embeddings as their similarity. The evaluation metrics is also the Spearman's rank correlation coefficient  $\rho$ between the average human rating and the cosine similarity scores given by our approach. 

\begin{table*}
\centering
\begin{tabular}{|l|l|}
\hline
 \hline System&$\rho\times100$\\
  \hline
Huang et al. \cite{HuangEtAl2012} (0.99B)&65.7\\
Chen et al. \cite{Chen2014AUM} (1B)+WordNet& 68.9\\
Tian et al.  \cite{Tian2014APM} (0.99B)&65.4\\
Neelakantan et al. \cite{Neelakantan2014EfficientNE}  (0.99B)&69.3\\
Li et al. \cite{Li2015DoME} (120B)& \textbf{69.7}\\
Liu et al. \cite{Liu2015TopicalWE} (0.99B) & 68.1\\
Wu and Giles \cite{wu2015sense}&66.4\\
Iacobacci et al. \cite{Iacobacci2015SensEmbedLS} &62.4\\\hline
SG (100B)&64.4\\
SG (100B)+Def (Word) (vanilla RNN)&66.2(+1.8)*\\
SG (100B)+Def (Word)+Def (Sense) (vanilla RNN)&66.8(+0.6)**\\
SG (100B)+Def (Word) (LSTM)&68.9(+4.5)* \\
SG (100B)+Def (Word)+Def (Sense) (LSTM)&69.5(+0.6)**\\
SG (100B)+Def (Word) (GRU)&69.1(+4.7)*\\
SG (100B)+Def (Word)+Def (Sense) (GRU)&\textbf{69.5}(+0.4)**\\
\hline
\end{tabular}
\caption{Performances for our system and other proposed approaches on SCWS dataset.  }
\label{contextwordsimilarity}
\end{table*}

Table  \ref{contextwordsimilarity} shows our results compared to previous approaches. Besides the models we have mentioned, Chen et al. \cite{Chen2014AUM} use WordNet to acquire number of senses of words  and use definitions just to initialize sense embeddings and then train sense embeddings on a corpus processed with WSD model.  Tian et al.  \cite{Tian2014APM} model word polysemy from a probabilistic perspective and combine it with Skip-Gram \cite{Mikolov2013DistributedRO} model. Liu et al. \cite{Liu2015TopicalWE} incorporate topic models into word sense embedding learning. Li et al. \cite{Li2015DoME} use Chinese Restaurant Processes to determine the sense of a word and learn the sense embeddings jointly.

As can be seen, the improvements from each training step of our approach are in accordance with the results in WordSim-353 evaluation. LSTM and GRU also present much more improvements than vanilla RNN. Our proposed approach present high overall performance on both word similarity tasks. That illustrates the word sense embeddings indeed capture the semantics of word senses. Strictly speaking, the comparison between different approaches are not totally fair because the resources different approaches use are different. 

\subsubsection{Word Sense Disambiguation Evaluation}

We also apply our word sense embeddings in a word sense disambiguation task to show the word sense embeddings capture the differences between senses of a word. In Semeval-2007 coarse-grained all-words WSD task \cite{Navigli2007SemEval2007T0}, WordNet is used as the word sense inventory. But the evaluation of word sense disambiguation result is on a coarser-grained version of the WordNet sense inventory and those word senses which are hard to disambiguate even for human are clustered into one class. The version of WordNet used in this task is 2.1, but we learn our word sense embeddings with WordNet 3.0. So we use the sense map\footnote{https://wordnet.princeton.edu/man/sensemap.5WN.html} between the two versions provided by the developers to address this issue.  To compare the effectiveness of our word sense embedding on this task with previous work, following Chen et al. \cite{Chen2014AUM}, we still adopt the S2C strategy we described in Section \ref{lastphase} to disambiguate word sense. We also show the result produced by randomly choosing the sense of words according to \cite{Chen2014AUM}.

\begin{table*}
\centering
\begin{tabular}{|l|l|}
\hline
 \hline System&F1\\
  \hline
Random&62.7\\
Chen et al. \cite{Chen2014AUM} (1B)+WordNet&75.8\\
\hline
SG (100B)+Def (Word) (vanilla RNN)&69.5\\
SG (100B)+Def (Word)+Def (Sense) (vanilla RNN)&70.3(+0.8)**\\
SG (100B)+Def (Word) (LSTM)&75.6 \\
SG (100B)+Def (Word)+Def (Sense) (LSTM)&\textbf{76.4}(+0.8)**\\
SG (100B)+Def (Word) (GRU)&75.7\\
SG (100B)+Def (Word)+Def (Sense) (GRU)&\textbf{76.3}(+0.6)**\\
\hline
\end{tabular}
\caption{Performances on Semeval-2007 coarse-grained all-words WSD task.  }
\label{wsd}
\end{table*}

The results are shown in Table \ref{wsd}. After we train our model and word sense embeddings using sense embeddings to represent words in definitions, our approach outperforms Chen et al. \cite{Chen2014AUM} on this task. It illustrates that our sense embeddings can actually distinguish different senses of a word and our approach can actually learn the semantics of senses from definitions.

\section{Related Work}
\label{relate}

Early word embedding learning approaches learn one embedding per word. Skip-gram \cite{Mikolov2013DistributedRO} and Glove \cite{Pennington2014GloveGV} are the most prevalent models of this kind. Both of them use context information extracted from an unannotated corpus to learn word embeddings.

Since  one embedding for each word sense are suggested to be better than a single embedding for a word, many word sense embedding learning approaches have been proposed \cite{Reisinger2010MultiPrototypeVM,HuangEtAl2012,Chen2014AUM,Neelakantan2014EfficientNE,Tian2014APM,Iacobacci2015SensEmbedLS,Liu2015TopicalWE,wu2015sense,Li2015DoME}. Researchers tend to extend Skip-gram and Glove models to learn sense embeddings with WSI or WSD as a preliminary. Reisinger and Mooney \cite{Reisinger2010MultiPrototypeVM} propose to cluster the contexts of each word into groups and make each cluster a distinct prototype vector. Huang et al. \cite{HuangEtAl2012} determine the sense of a word by clustering the contexts and then apply it to neural language model with global context. Guo et al. \cite{Guo2014LearningSW} propose to use parallel data for WSI and learning word sense embeddings. Neelakantan et al. \cite{Neelakantan2014EfficientNE} extend Skip-gram \cite{Mikolov2013DistributedRO} to a model which jointly performs word sense discrimination and embedding learning.  Liu et al. \cite{Liu2015TopicalWE} associate words with topics and then extend Skip-gram \cite{Mikolov2013DistributedRO} to learn sense and topic embeddings. Wu and Giles \cite{wu2015sense} propose to use Wikipedia concepts to cluster word senses and to learn sense-specific embeddings of words. Li et al. \cite{Li2015DoME} use Chinese Restaurant Processes to determine the sense of a word and learn the sense embedding jointly. Iacobacci et al. \cite{Iacobacci2015SensEmbedLS} use BabelNet \cite{Navigli2012BabelNetTA} as the word sense inventory and opt for Babelfy \cite{Moro2014EntityLM} to perform WSD on Wikipedia\footnote{http://dumps.wikimedia.org/enwiki/}. Then they train word sense embeddings using CBOW architecture \cite{Mikolov2013EfficientEO} on the processed corpus. Chen et al.\cite{Chen2014AUM} use WordNet as its lexical ontology to acquire numbers of word senses and use the average word embedding of words chosen from definitions as the initialization of sense embeddings. And then they do WSD on a corpus and train sense embeddings with a variant of Skip-gram on the corpus. Both of our approaches use words in definitions to initialize word sense embeddings, but after that their training still concentrates on the corpus while we train our model and word sense embeddings with definitions. The disadvantage to use a corpus processed by WSD or WSI may come from the unreliability of the processing results and since a corpus for embedding training is usually much larger in scale than the summation of all the definitions to get satisfied result, their approach inevitably consumes much more time on WSD and training.

\section{Conclusion}
\label{conclusion}
In this paper, we propose to use RNN-based models to learn word sense embeddings from sense definitions. Our approach produces an effective natural language understanding model and word sense embeddings of high quality. Comparing with previous work training word sense embeddings on a corpus, our approach is less time-consuming and better for rare word senses. Experimental results show our word sense embeddings are of high quality.  

\section*{Acknowledgments}
This work is supported by National Key Basic Research Program of China under Grant No.2014CB340504 and National Natural Science Foundation of China under Grant No.61273318. The Corresponding author of this paper is Baobao Chang.

\bibliographystyle{splncs03}
\bibliography{nlpcc}

\begin{thebibliography}{10}
\providecommand{\url}[1]{\texttt{#1}}
\providecommand{\urlprefix}{URL }

\bibitem{Bahdanau2014NeuralMT}
Bahdanau, D., Cho, K., Bengio, Y.: Neural machine translation by jointly
  learning to align and translate. CoRR  abs/1409.0473 (2014)

\bibitem{Chen2014AUM}
Chen, X., Liu, Z., Sun, M.: A unified model for word sense representation and
  disambiguation. In: EMNLP (2014)

\bibitem{Chung2014EmpiricalEO}
Chung, J., Çaglar G{\"u}lçehre, Cho, K., Bengio, Y.: Empirical evaluation of
  gated recurrent neural networks on sequence modeling. CoRR  abs/1412.3555
  (2014)

\bibitem{Finkelstein2001PlacingSI}
Finkelstein, L., Gabrilovich, E., Matias, Y., Rivlin, E., Solan, Z., Wolfman,
  G., Ruppin, E.: Placing search in context: the concept revisited. ACM Trans.
  Inf. Syst.  20,  116--131 (2001)

\bibitem{Guo2014LearningSW}
Guo, J., Che, W., Wang, H., Liu, T.: Learning sense-specific word embeddings by
  exploiting bilingual resources. In: COLING (2014)

\bibitem{Hermann2015TeachingMT}
Hermann, K.M., s~Kocisk\'{y}, T., Grefenstette, E., Espeholt, L., Kay, W.,
  Suleyman, M., Blunsom, P.: Teaching machines to read and comprehend. CoRR
  abs/1506.03340 (2015)

\bibitem{Hochreiter1997LongSM}
Hochreiter, S., Schmidhuber, J.: Long short-term memory. Neural Computation  9,
   1735--1780 (1997)

\bibitem{HuangEtAl2012}
Huang, E.H., Socher, R., Manning, C.D., Ng, A.Y.: Improving word
  representations via global context and multiple word prototypes. In: Annual
  Meeting of the Association for Computational Linguistics (ACL) (2012)

\bibitem{Iacobacci2015SensEmbedLS}
Iacobacci, I., Pilehvar, M.T., Navigli, R.: Sensembed: Learning sense
  embeddings for word and relational similarity. In: ACL (2015)

\bibitem{Kalchbrenner2013RecurrentCT}
Kalchbrenner, N., Blunsom, P.: Recurrent continuous translation models. In:
  EMNLP (2013)

\bibitem{Li2015DoME}
Li, J., Jurafsky, D.: Do multi-sense embeddings improve natural language
  understanding? In: EMNLP (2015)

\bibitem{Liu2015TopicalWE}
Liu, Y., Liu, Z., Chua, T.S., Sun, M.: Topical word embeddings. In: AAAI (2015)

\bibitem{Mikolov2013EfficientEO}
Mikolov, T., Chen, K., Corrado, G., Dean, J.: Efficient estimation of word
  representations in vector space. CoRR  abs/1301.3781 (2013)

\bibitem{Mikolov2013DistributedRO}
Mikolov, T., Sutskever, I., Chen, K., Corrado, G., Dean, J.: Distributed
  representations of words and phrases and their compositionality. CoRR
  abs/1310.4546 (2013)

\bibitem{Miller1992WordNetAL}
Miller, G.A.: Wordnet: A lexical database for english. Commun. ACM  38,  39--41
  (1992)

\bibitem{Moro2014EntityLM}
Moro, A., Raganato, A., Navigli, R.: Entity linking meets word sense
  disambiguation: a unified approach. TACL  2,  231--244 (2014)

\bibitem{Navigli2007SemEval2007T0}
Navigli, R., Litkowski, K.C., Hargraves, O.: Semeval-2007 task 07:
  Coarse-grained english all-words task (2007)

\bibitem{Navigli2012BabelNetTA}
Navigli, R., Ponzetto, S.P.: Babelnet: The automatic construction, evaluation
  and application of a wide-coverage multilingual semantic network. Artif.
  Intell.  193,  217--250 (2012)

\bibitem{Neelakantan2014EfficientNE}
Neelakantan, A., Shankar, J., Passos, A., McCallum, A.: Efficient
  non-parametric estimation of multiple embeddings per word in vector space.
  In: EMNLP (2014)

\bibitem{Pennington2014GloveGV}
Pennington, J., Socher, R., Manning, C.D.: Glove: Global vectors for word
  representation. In: EMNLP (2014)

\bibitem{Reisinger2010MultiPrototypeVM}
Reisinger, J., Mooney, R.J.: Multi-prototype vector-space models of word
  meaning. In: NAACL (2010)

\bibitem{Rocktschel2015ReasoningAE}
Rockt\"{a}schel, T., Grefenstette, E., Hermann, K.M., s~Kocisk\'{y}, T.,
  Blunsom, P.: Reasoning about entailment with neural attention. CoRR
  abs/1509.06664 (2015)

\bibitem{Tian2014APM}
Tian, F., Dai, H., Bian, J., Gao, B., Zhang, R., Chen, E., Liu, T.Y.: A
  probabilistic model for learning multi-prototype word embeddings. In: COLING
  (2014)

\bibitem{wu2015sense}
Wu, Z., Giles, C.L.: Sense-aware semantic analysis: A multi-prototype word
  representation model using wikipedia (2015)

\bibitem{Zeiler2012ADADELTAAA}
Zeiler, M.D.: Adadelta: An adaptive learning rate method. CoRR  abs/1212.5701
  (2012)

\end{thebibliography}

\end{document}